\documentclass{IOS-Book-Article}

\usepackage{mathptmx}
\usepackage{graphicx}
\usepackage{booktabs}
\usepackage{subfig}

\def\hb{\hbox to 10.7 cm{}}

\begin{document}

\pagestyle{headings}
\def\thepage{}

\begin{frontmatter}              

\title{The Role of Normware\\in Trustworthy and Explainable AI}

\markboth{}{December 2018\hb}

\author[A]{\fnms{Giovanni} \snm{Sileno}%
\thanks{Corresponding Author: 
g.sileno@uva.nl.}},
\author[B]{\fnms{Alexander} \snm{Boer}}
and
\author[B,A,C]{\fnms{Tom} \snm{van Engers}}

\runningauthor{G. Sileno et al.}
\address[A]{Informatics Institute, University of Amsterdam, Netherlands}
\address[B]{Leibniz Institute, University of Amsterdam/TNO, Netherlands}
\address[C]{Institute for Advanced Study (IAS), University of Amsterdam, Netherlands}

\begin{abstract}
For being potentially destructive, in practice incomprehensible and for the most unintelligible, contemporary technology is setting high challenges on our society. New conception methods are urgently required. Reorganizing ideas and discussions presented in AI and related fields, this position paper aims to highlight the importance of \emph{normware}--that is, computational artifacts specifying norms--with respect to these issues, and argues for its irreducibility with respect to software by making explicit its neglected ecological dimension in the decision-making cycle.
\end{abstract}

\end{frontmatter}
\markboth{December 2018\hb}{December 2018\hb}

\section{Introduction}
With the (supposedly) near advent of autonomous artificial entities, or other forms of distributed automatic decision-making in all human activities, legitimate concerns are raised about the increase of risks of consequences that were not planned, intended or desired at design time. 
At par with this, most contemporary decision making applications, in particular those based on statistical machine learning, rely on computational methods that for the most are not transparent, i.e. whose internal workings are opaque and/or too complex, nor they are capable to furnish explanations
on why a certain decision has been taken. 
In the last few years several dedicated calls about these critical issues have been started by public and private actors.

Within this wave of studies, this position paper aims to set novel arguments for explicitly considering \textit{normware} in the conception of artificial devices.\footnote{Here ``norm'' is used in a general sense, as in \textit{normative} (shared drivers) and in \textit{normal} (shared expectations).} Normware can be perceived from two complementing perspectives. On the one hand, normware consists of \textit{computational artifacts specifying norms}. The basic intuition is that, if used to circumscribe and direct practical and epistemic actions by placing adequate checks and balances in the decision-making cycle, normware offers a mean to deal with the problems of trustworthiness and explainability. However, the effort of transforming regulations and knowledge into computational artifacts, and/or to (partially) automatize judgment has a long history, and it is currently renewed by studies in the domain of artificial ethics and responsible AI. Recent developments in RegTech and FinTech share also similar motivations. 
In this context, one might reasonably doubt whether the introduction of the ``normware'' concept could bring anything new, seen that studies and practices have started exploring this path since a long time already. Yet, the paper shows there is still a conceptual grey area concerning the functions of norms as computational artifacts, in particular at architectural level, where the aforementioned checks and balances can play a role. This uncovers the other perspective on normware, as the \textit{ecology of normative components} \textit{that guide} (but do not and cannot control) \textit{the behavior of a system}.

\section{Problems and causes of problems}

With humans less and less in the loop, conception methods are required to put in place protections and possibly to set up remedy or repair actions against the occurrence of unintended consequences of automated decisions. Problems might arise already at epistemic level, when some proposed conclusion would be not acceptable by relevant human standards. This section explores the possible causes of such scenarios. 



\subsection{Unintended consequences because of wrong conception}
In software development, two types of cause of unintended consequences can be recognized, both going down to a failure in the conception of the device: \textit{implementation faults}: that is, actual errors in the development, as ``bugs'' in software; or 
\textit{design faults}, due to a contextually limited conception because certain possible scenarios were not considered at design time. 
Implementation faults might be prevented via formal verification, but only to the extent in which the constraints to be verified are correctly specified. Design faults are instead inescapable, as the ``real'' environment will necessarily be always richer than any model of it. Working solutions are feasible only in the case in which the environment's stability is sufficient to produce only a few, minor exceptions to the system model used at design time. However, when the environment is a human social system this assumption often becomes untenable, for the diversity and variability of behaviors that can possibly be encountered. Furthermore, because any cyber-physical chain will have eventually to interact with humans, 
infrastructures deemed to operate merely at technical level will eventually derive their semantics from the social interactions they enable, and for this reason they could always be used for purposes unintended at design time.\footnote{To size up today's threats for the reader, consider for instance \textit{blockchain} technology, marketed in the last few years as the most secure solution to almost anything: only in in 2017 at least half a billion euros were lost through software bugs, wallet hacks and fraudulent actions \cite{De2017}.}



\subsection{Unintended consequences because of improvident induction}
The second type of cause is more subtle, but not less dangerous, and concerns specifically applications based on statistical machine learning.
Consider for instance the case of the software used in the US for predicting recidivism for petty criminals, proven to biased against Afro-Americans (2016) \cite{Angwin2016}. Even if the statistical bias resulting from the training data correctly describes the frequency of a certain event in reality, when it is applied for prediction on an individual instance, this \textit{description} is implicitly transformed into a \textit{mechanism}, as it is interpreted as a behavioural predisposition (see also \cite{Amaranto2017}). But, by introducing in the judiciary activity a bias against members of a certain community there would be harsher penalties, more segregation, then decreasing opportunities to a fair access to the economic cycle for that community, which eventually would reinforce the statistical source of the bias, producing a ``self-fulfilling prophecy'' effect. 

The core problem here is the integration of statistical inference in judgment. In tribunal, this question relates to the role of circumstantial evidence. If very few people would argue against DNA, many will reject arguments based on the country of origin, gender, ethnicity or wealth as improper profiling. Where lies the difference? Both integrations introduce probabilistic measures, but the second clearly undermines essential rights guaranteed by the law (e.g. equality before the law) promoting neutrality and fairness, because individuals are judged as members of certain aggregates and not as individuals. The ``improper" qualification is actually a consequence of these rights.\footnote{As a counter-example, consider a recommendation system using factors related to ethnicity (plausibly not declared but resulting from data-mining techniques) for suggesting dress colors; very few people would argue against it. 
}

\subsection{Unacceptable conclusions because of improvident induction}
The ``improvident'' qualification to an inductive inference might be given already before taking into account the practical consequences of its acceptation. Consider a biomedical device predicting whether the patient has appendicitis by analyzing all kind of personal and environmental data. We could accept easily a conclusion based on factors as the presence of fever, abdominal pain, etc. 
but we would raise some doubt if it would be based e.g. on the length of the little toe, or even more, on the fact that it is raining. Intuitively, to accept an inference we need at least to suspect some mechanism deemed relevant for that decision-making context; consequently, an expert would reject the conclusion when no mechanism can be reasonably imagined linking that factor with the conclusion.

Statistical inference is also vulnerable to counter-intuitive results, as with the famous Simpson's paradox \cite{Pearl2014}, 
a phenomena whereby the association between a pair of variables X and Y reverses sign when adding a third variable Z. For instance, the association between gender (X) and being hired (Y) by a university may for instance reverse on knowing the hiring department (Z), e.g. women result favoured at department level, but men result so at university level.\footnote{E.g. the mathematics department has two positions, and one woman and ten men as applicants. One man and one woman are hired. The department apparently favours women (1/1) over men (1/10). The sociology department has one position, and a hundred women and one man apply. A woman is hired. The sociology department favours women over men (1/100 vs. 0/1) as well, but the university as a whole favours men over women (1/11 vs. 2/101).} 
This is initially surprising, but makes sense when considering that some proposed underlying mechanisms allow for this reversal and others do not \cite{Pearl2014}. Whether one should consider either X or X and Z in conjunction acceptable as a relevant factor for inferring Y is a matter of choosing a mechanism. 


\paragraph{Expert vs non-expert informational asymmetry} The interplay with some expert standard plays a fundamental role in determining what is acceptable or not. 
In these days, technology seems to naturally push us towards non-expert positions, because of the indirect persuasive effect of the complicated mathematics involved in machine learning and of the use of computers (which don't ``make mistakes''), but this is happening not without resistance. It is not fortuitous that research in the healthcare sector is striving to implement \textit{explainable AI} for evidence-based medicine: doctors want to understand why a certain conclusion is made by a certain device before accepting its response, because they have responsibility with respect to their patients (not only legal, but also moral). This consideration clearly applies as well on legal judgment.\footnote{As a counter-example, consider face detection: we usually do not ask the system why it recognizes or does not recognize a person, unless we are the developers of that application. 
}

\section{What we have, what we do not have}

On the light of what exposed above, the main driver of \textit{explainable AI} can be identified as \textit{satisfying reasonable requirements of expertise, including rejecting unacceptable arguments}. 
On the other hand, \textit{trustworthy AI} can be associated to the requirement of not falling into ``paperclip maximizer'' scenarios \cite{Bostrom2003}, i.e. of \textit{not taking ``wrong'' decisions, of performing ``wrong'' actions, wrong because having disastrous impact for a certain reasonable standard}. These minimal definitions implies that, in contrast to current literature in Responsible AI (e.g. \cite{Dignum2017}), trustworthiness and explainability might be separated properties. In effect, following common sense, trustworthiness does not necessarily require explainability; for instance, small children trust their parents without requiring explanations. On the other hand, explainability does not require trustworthiness, for instance criminals might explain their conduct plainly.\footnote{However, one might require trustworthiness \textit{in} providing explanations, in the sense of establishing an alignment between the actual decision-making process and its justification. 
This \textit{transparency} requirement is necessary to unmask hidden agendas, but it is arguable whether it has always positive effects if rigorously applied.
}

At this point, we need to understand what defines an expertise to be ``reasonable'', what qualifies an argument as ``unacceptable'' and how to denote actions and consequences/situations as ``undesirable''. 

Experts are expected to refer to shared conceptualizations established in their own domain while forming explanations. An argument is then \textit{epistemically acceptable} in the moment in which it is valid and coherent with this expert knowledge (usually integrated with some common-sense). Additionally, in order to have \textit{practical acceptability}, one has to check whether when used to make predictions, it does not cause undesirable effects or reinforces undesirable mechanisms. Interestingly, this connects with the requirement concerning the specification of desirable/undesirable actions and situations.

At superficial level, all these demands mirror research subjects traditionally investigated in knowledge engineering and related disciplines.\footnote{Shared conceptualizations can be--to a certain extent--captured by formal ontologies or other expert-systems knowledge artifacts, including probabilistic representations (e.g. Bayesian networks). The specification of desires and preferences is the basis for requirement engineering, for agent-based programming (belief-desires or beliefs-desires-intentions architectures), has associations with logic and constraint programming paradigms (seeing queries as positive desires, constraints as negative desires, etc.) but also with deontic logic and other logics of desires and preferences, and with notations representing user preferences (CP-nets, GAI networks).} 
The important message here is that all representational models capture some aspect of epistemic and practical commitments, and for this reason they could provide in principle (or at least suggest) some of the means necessary to specify normware. In contrast, none of the associated methods provides--nor aims to provide--a \textit{general theory} on how normware would operate in decision-making (and not merely as logic component of an algorithm); and in particular on how it would relate to sub-symbolic modules, in \textit{hybrid reasoning} environments. To address this question we now focus our investigation on the core of decision-making.

\section{The role of normware}


\subsection{Defining problems sufficiently well}
Traditionally, engineering is concerned with the conception of devices implementing certain functions, defined within a certain operational context. For instance, we need houses as shelters to live in, car for moving, phones to communicate. The primary driver in design is \textit{satisfying given needs}. Because in practical problems, when a solution is possible, many alternatives are possible too, a second concern of engineering is to conceive the best device, according to certain criteria. 
The secondary driver is then \textit{optimality over preferences}, e.g. by maximization of rewards. 

Needs, preferences and the resources available for the device or process development are the basic components of a \textit{well-defined problem}, input to design or planning problem-solving methods \cite{Breuker1994}. The acquisition and processing of these elements is however not without critical aspects. First, enumerating all needs, preferences and conditions at stake is difficult because many situational and volitional defaults are taken as granted, very often in an unconscious way. 
Second, a focused closure, placing some of the collected elements in the foreground and the remaining in background, is inevitably necessary to balance with limited cognitive resources. 

\subsection{Dividing tactical and strategic concerns}
For the sake of the argument, suppose that our goal is fishing, and our established reward increases with the quantity of fish, and decreases with the effort. The problem being well-defined, we can delegate it to an automatic planner, which, to our surprise, finds that the best solution is fishing with bombs. Clearly, this plan would entail undesired second-order effects as the destruction of ecological cycles, resulting in the longer run in a reduction of the quantity of fish; for this reason, we need to write down additional constraints to strip out such blatantly dangerous solution. The full decision-making process covered thus two steps, taking two different standpoints--that we may call \textit{tactical} and \textit{strategic}--with the second able to evaluate outputs from the first. Integrating an operational step to capture perception and actual execution, we obtain the scheme in Fig.~1.

\begin{figure}
    \centering
    \scalebox{0.25}{\includegraphics{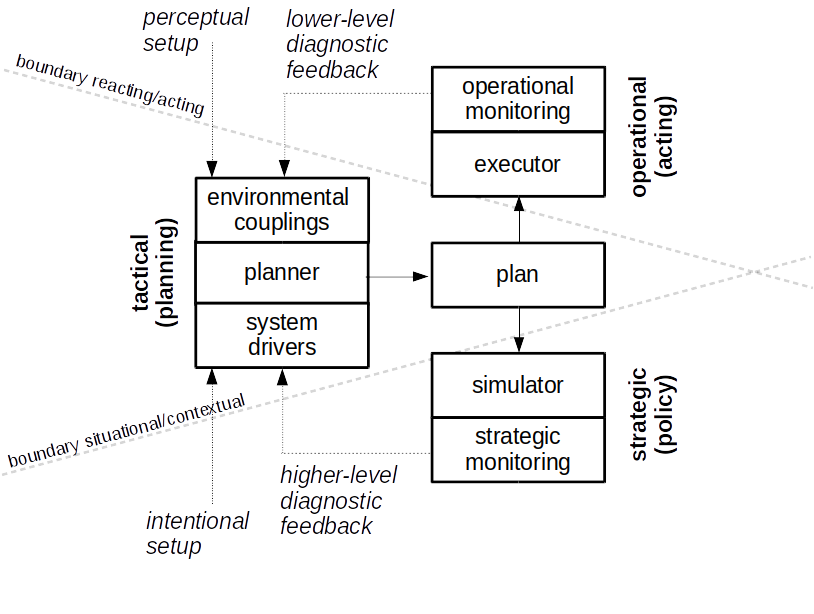}}
    \vspace{-15pt}
    \caption{Tactical, strategic and operational levels in practical decision-making}
\end{figure}


The proposed diagram presents three levels, in affinity with models used to categorize activities in organizations (e.g. operations, product/service development and policy  \cite{Boer2011a}). The \textit{tactical level} is centered around the planner, which, exploiting the available knowledge (specifying expectations, abilities and susceptibilities, i.e. the \textit{system-environment couplings}) and the goals and constraints given in input (the \textit{system drivers}) produces a plan. The \textit{strategic level} sets up the initial goals for the tactical layer but also rechecks the plan by predicting its consequences through the knowledge available from a strategic standpoint (generally with coarser granularity but wider spatio-temporal range). In case a conflict is observed, a higher-level diagnostic feedback refines the system drivers. The \textit{operational level} feeds the planner with perceptual information about the current situation; once the plan is received, this is executed and if there is a conflict between actual and expected outcome, a lower-level diagnostic feedback is sent back to the tactical level (stating e.g. that a certain module is not working properly or that the environment response follows a different pattern). Functionally, this architecture enables a closure on the requirements and on the information fed at tactical level to guarantee that the optimization performed by the planner is feasible.



\subsection{Machine learning as decision-making}
The three levels illustrated above might be assigned to different agents. In our fishing scenario, the tactical level is covered by a planning software, the strategic level by a human user. Other configurations are possible. For instance, for their reactivity, machine learning feed-forward networks are possible candidates for the operational level, although the interface with the other levels needs to be investigated.

Let us consider supervised machine learning. In operational settings, a ML black-box is a feed-forward network of parameter-controlled computational components implementing some function; for instance, in case of a classifier, it takes as input some object and returns a prediction of its class. In training, the parameters are adapted by means of some ML algorithm taking into account some feedback, e.g. the error between the actual outcome and the desired outcome expressed by an \textit{oracle}. An implicit assumption of this architecture is that, because the oracle is deemed to be correct in its prediction, it has access to more knowledge relevant to interpret the input than the black-box. 
Interpreting the whole process through the decision-making model presented above, the following correspondences can be observed:

\vspace{-4pt}
\begin{table}[h]
\scalebox{1}{
\noindent \begin{tabular}{ll} 
\emph{ML black-box component}  & \emph{function in decision-making cycle} \\ \midrule
data-flow computational network & executor \\
parameters distributed along the network & plan \\                 ML method enabling adaptation of parameters & planner \\
adaptation against error & lower-level diagnostic feedback \\
oracle & intentional setup and expected outcome (linked to plan) \\ 
\end{tabular}
}
\end{table}
\noindent These analogies unveil that ML black-boxes feature only a partial strategic level, as it does not feature a higher-level diagnostic feedback. This offers an explaination of why ML is particularly vulnerable to explainable and trustworthy AI issues. 



\subsection{Distributing computation to a social network}
The adaptation mathematically controlled by the ML method (e.g. via gradient descent) can be reinterpreted in evolutionary terms as a competition for scarce computational resources. Rather than one black-box modifying itself, we can consider the presence of a multitude of different non-adaptive black-boxes, covering several configurations of parameters. For each learning step, the oracle sets the means to select the best performing black box, for which access to computational resources for future predictions will be granted as a reward. This metaphor enables us to think different configurations in the spirit of genetic algorithms. For instance, a threshold on performance might be specified to maintain a set of black-boxes for each step (only the best one would produce the actual output though); rather than starting with all possible configurations, at each step a set of mutations of current blackboxes could be added to enable evolution; 
etc. Leaving the specification of this infrastructure to future study, we focus on the architectural issue. 

\begin{figure}
    \centering
    \scalebox{0.25}{\includegraphics{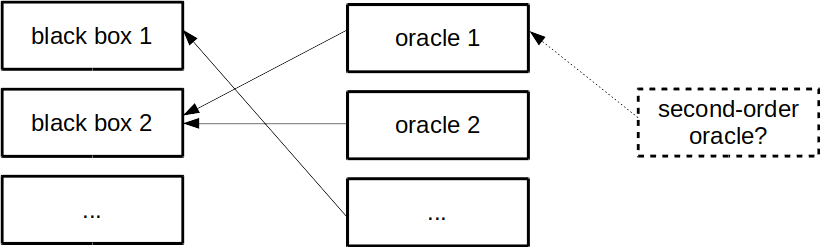}}
    \caption{Distributed rewards scheme}
\end{figure}


In the decision-making model presented above, the presence of a higher-level diagnostic feedback implies that also the system drivers should pass from some selection mechanism. This consideration is also sensible from an evolutionary point of view: if the decision-making of the oracle has to be embedded into the autonomous system, it will use computational resources as well and then it should compete for them as black-boxes do. Thus, we need to add to the previous network a second multitude, of oracles this time, and a second-order oracle that guides their selection (Fig. 2).\footnote{ 
As a concrete example, consider a well-known success story of contemporary AI systems: IBM Watson \cite{Ferrucci2010}. Watson has, amongst other applications, shown great competence at winning the game show Jeopardy. 
Architecturally, it behaves like a coalition of intelligent question-answering (QA) components that compete against each other to produce the best answer to a question, and in the process of doing so it acquires feedback on its performance and becomes better at selecting the best competitor. 
Reinterpreting this architecture through the model in Fig. 2, the initial question acts as an input triggering the decision-making cycle, the system has then to guess what the question demands (by selecting the \textit{oracles}) and what the answers to these demands might be (\textit{black-boxes}); the final response given by the jury (social environment acting as \textit{second-order oracle}) enables the reinforcement of correct alignments between the components.}

\subsection{Setting up checks and balances}
The analysis above presented the essence of the \textit{ecological perspective} on normware: an internal system of positive and negative rewards that may be used to maintain a coalition of specialized agents proven to be intelligent in specific operational niches (computational/symbolic or physical). Applying this architecture, we can consider a wider range of applications. 

\begin{figure}
\hspace{-30pt}
    \centering
    \subfloat[\label{fig:test1}]{\makebox[0.45\textwidth]{\scalebox{0.22}{\includegraphics{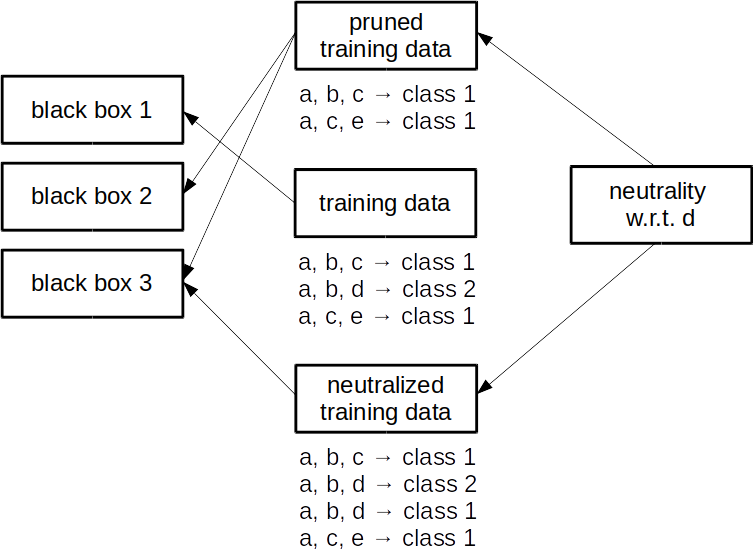}}}} \qquad\qquad
    \subfloat[\label{fig:test2}]{\makebox[0.45\textwidth]{\scalebox{0.22}{\includegraphics{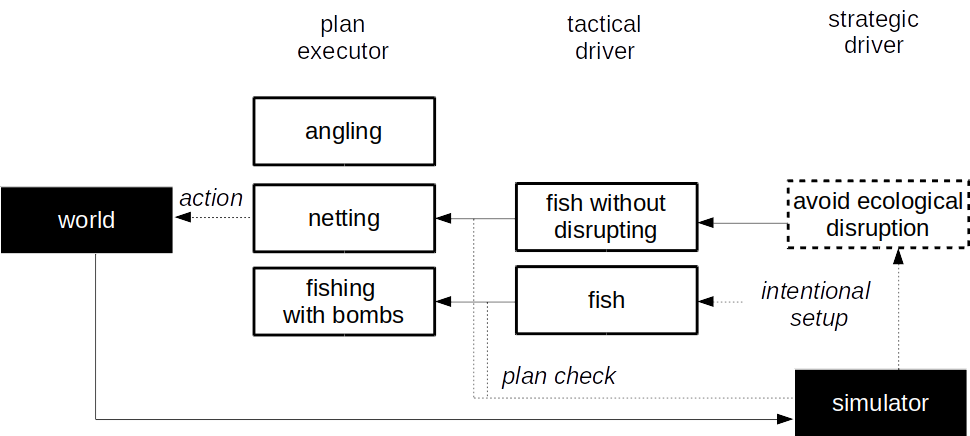}}}}
    \caption{Two examples of applications of normware: neutrality constraints modifying the drivers derived from ML training (a), strategic protection to unintended consequences (b).}
\end{figure}


\paragraph{Neutrality with respect of a certain factor}
Suppose that we need to build up a predictor based on training data, but that this should also neutral with respect to a certain factor. The source training data (acting as an oracle) would reward a black-box potentially vulnerable to statistical biases for that factor. A second-order oracle specified by a neutrality constraint would instead promote mutations of the source training data that satisfy it, e.g. pruning the data introducing the bias, or adding additional data to neutralize it (Fig.~3a).  



\paragraph{Protection against unintended consequences}
The fishing case 
adds to the previous case the simulator component and the interface with the physical world (Fig. 3b). The decision making cycle starts from the intentional setup: the intent to fish. The optimization is made at the tactical level, selecting with rewards the fishing with bombs method. The selected plan is preventively checked in a virtual environment through the simulator. In this case there is conflict, so the strategic driver rewards an alternative tactical driver with an additional constraint. The constraint results in a different plan which passes the strategic check and it is then executed. Finally, the fact that plan works and does not result in ecological destruction rewards the strategic driver and knowledge.

\paragraph{Acceptable factors and mechanisms}
The biomedical device predicting appendicitis provides a test case for explainability. At functional level the system might seen as consisting of two independent sub-systems: one predicting the conclusion and another one deciding how to justify that conclusion to the user. The two sub-systems could be in principle completely separated, save for the functional dependence. The input of the justification sub-system is the conclusion of the prediction sub-system together with the perceptual input and the desired output is a train of reasoning which is compatible to the expert knowledge model. The diagram would be very similar to the fishing case, but instead of checking the plan through the simulator, here the explanation would be tested against the expert knowledge. Interestingly, both cases can be thought as forms of \textit{alignment checking}. Conflicts will direct modifications of the tactical modules producing explanations towards providing an acceptable one. 


\section{Perspectives}

This contribution aims to present novel arguments for bringing to the foreground the role of norms, and then of their computational counterpart normware, for dealing with requirements of trustworthiness and explainability in computational systems. Looking at the current trends, researches developed in the area of machine learning usually overlook or only implicitly consider this level of abstraction. On the other hand, the term ``normware'', in continuity to hardware and software, has been decided to explicitly refer to implementation aspects, usually put aside in higher-level contributions as those currently presented in artificial ethics and responsible AI. 

A key tenet of the paper is that normware is not so much a matter of knowledge representation, as it is a matter of understanding how the knowledge artifact is placed within the decision-making cycle of the autonomous system, ecologically coexisting with other components, including other normware. For this reason, the proposed architecture presents natural analogies with paradigms as \textit{epistemic} and \textit{legal pluralism}. A problem domain useful to explain this change of perspective is \textit{ontological alignment}: researchers in formal ontologies attempt to unify or connect ontologies at conceptual level, i.e. without considering the interactional niches which motivated the very existence of these ontologies (e.g. specific domain of applications, cf. \cite{Boer:03c}). The architecture presented here proposes an alternative approach. Depending on the input situation a certain ontology will be selected instead than another because it is expected to be more successful in interpreting it. Therefore, two concepts belonging to different ontologies might be aligned not because they share the underlying logical structure, but because they produce a similar function for that type of input situation (cf. \cite{Boer:03c}).

Going further, the neutrality of the architecture with respect to the representational model leaves open the possibility, as in IBM Watson, to have components which are not necessarily built using symbolic AI. Could we still talk about normware in this case? Although the processing may eventually not be based on symbolic methods, normware artifacts--the ones that are used at the interface between the \textit{guiding} agents (users, developers, society, etc.) and supposedly \textit{guided} agents (the autonomous artificial system)--rely on symbols, just as norms are communicated through language by humans. 


\end{document}